\documentclass[manuscript,acmsmall,review,screen]{acmart}

\usepackage[]{hyperref}
\usepackage[utf8]{inputenc}
\usepackage{kotex}
\usepackage{cleveref}
\usepackage{lipsum}
\usepackage{xspace}
\usepackage{listings}
\usepackage{multirow}
\usepackage{makecell}
\usepackage{tikz}
\newcommand*\circled[1]{\tikz[baseline=(char.base)]{
            \node[shape=circle,draw,inner sep=0.5pt] (char) {#1};}}

\usepackage{enumitem}
\setlist{leftmargin=18pt}

\usepackage[normalem]{ulem}

\usepackage[subrefformat=parens,labelformat=parens,caption=false,font=footnotesize]{subfig}

\newcommand{\XXX}{\texttt{Puzzle}\xspace}

\newif\ifshowcomments
\showcommentstrue

\ifshowcomments
  \newcommand{\ys}[1]{[\textcolor{blue}{\textit{YS: #1}}]}
  \newcommand{\note}[1]{\textcolor{blue}{#1}}
  \newcommand{\duseok}[1]{[\textcolor{OrangeRed2}{\textit{DS: #1}}]}
  \newcommand{\jh}[1]{[\textcolor{Green}{\textit{JH: #1}}]}
\else
  \newcommand{\ys}[1]{}
  \newcommand{\note}[1]{}
  \newcommand{\duseok}[1]{}
  \newcommand{\jh}[1]{}
\fi

\begin{document}

\title{\XXX: Scheduling Multiple Deep Learning Models on Mobile Device with Heterogeneous Processors}

\author{Duseok Kang}
\email{duseok@qti.qualcomm.com}
\orcid{0000-0003-4985-0789}
\author{Yunseong Lee}
\email{yunseong@qti.qualcomm.com}
\author{Junghoon Kim}
\email{juk@qti.qualcomm.com}
\affiliation{%
  \institution{Qualcomm AI Research}
  \city{Seoul}
  \country{Korea}
}\thanks{Qualcomm AI Research is an initiative of Qualcomm Technologies, Inc.}

\begin{abstract}
  As deep learning models are increasingly deployed on mobile devices, modern mobile devices incorporate deep learning-specific accelerators to handle the growing computational demands, thus increasing their hardware heterogeneity. However, existing works on scheduling deep learning workloads across these processors have significant limitations: most studies focus on single-model scenarios rather than realistic multi-model scenarios, overlook performance variations from different hardware/software configurations, and struggle with accurate execution time estimation. To address these challenges, we propose a novel genetic algorithm-based methodology for scheduling multiple deep learning networks on heterogeneous processors by partitioning the networks into multiple subgraphs. Our approach incorporates three different types of chromosomes for partition/mapping/priority exploration, and leverages device-in-the-loop profiling and evaluation for accurate execution time estimation. Based on this methodology, our system, \XXX, demonstrates superior performance in extensive evaluations with randomly generated scenarios involving nine state-of-the-art networks. The results demonstrate \XXX can support 3.7 and 2.2 times higher request frequency on average compared to the two heuristic baselines, NPU Only and Best Mapping, respectively, while satisfying the equivalent level of real-time requirements.

\end{abstract}

\maketitle 
\pagestyle{plain} 

\section{Introduction}
Deep Learning (DL) has achieved remarkable performance across various fields, leading to the widespread adoption of numerous and diverse DL models. As these models are deployed in diverse applications, the computational and memory demands of DL workloads have surged. This has prompted the integration of specialized Neural Processing Units (NPUs) tailored for DL workloads, offering more efficient processing than traditional hardware accelerators like CPUs and GPUs. Consequently, systems equipped with multiple heterogeneous processors have become common. This has spurred extensive research into methodologies that leverage these heterogeneous processors to handle DL workload effectively~\cite{kumar2022overflowing,spatio-temporal-sharing2022atc,silvano2024surveydeeplearninghardware,aurora2024li_optimizing_moe}.

\begin{table*}[t]
    \footnotesize
    \centering
    \caption{Comparison various works utilizing heterogeneous processors to accelerate DL networks~\label{tab:scheduling-comparison}}
    \begin{tabular}{lcccc}
        \toprule
        & \makecell[c]{\textbf{Partition} \\\textbf{Unit}} & \makecell{\textbf{Multiple} \\\textbf{Models}} & \makecell{\textbf{Search} \\\textbf{Space}} & \makecell{\textbf{Non-linearity of} \\\textbf{Exec. Time (\S~\ref{sec:non-linearity})}} \\
        \midrule
        \(\mu\)layer~\cite{mulayer}           & Layer & \(\times\)    & \texttt{P} + \texttt{BR}                                & \(\times\) \\
        CoDL~\cite{codl}                      & Layer & \(\times\)    & \texttt{P}                                              & \(\times\) \\
        Kang et al.~\cite{kang2020scheduling} & Graph & \(\triangle\) & \texttt{M}                                              & \(\times\) \\
        Band~\cite{band}                      & Graph & \(\bigcirc\)  & \texttt{M}                                              & \(\bigcirc\) \\
        Jeong et al.~\cite{jeong2022tensorrt} & Graph & \(\times\)    & \texttt{M}                                              & \(\bigcirc\) \\
        Collage~\cite{collage}                & Graph & \(\times\)    & \texttt{M} \(\times\) \texttt{BE}                       & \(\bigcirc\) \\
        OmniBoost~\cite{omniboost}            & Graph & \(\bigcirc\)  & \texttt{M}                                              & \(\times\) \\
        \midrule
        \textbf{\XXX} (ours)                  & Graph & \(\bigcirc\)  & \texttt{M} \(\times\) \texttt{T} \(\times\) \texttt{BE} & \(\bigcirc\) \\
        \bottomrule
        \multicolumn{5}{r}{\texttt{P}: Partition Ratio / \texttt{BR}: Branch Dist. / \texttt{M}: Mapping / \texttt{T}: Data Type / \texttt{BE}: Backend Impl.} \\
    \end{tabular}
\end{table*}

When it comes to platforms with limited resources, such as mobile devices, there is also a wide variety of research that focuses on utilizing heterogeneous processors~\cite{mulayer,codl,collage,kang2020scheduling,taco2021-edgeduler,jeong2022tensorrt,band,omniboost}. Unfortunately, these works often overlook critical aspects of DL workloads on mobile devices.
\Cref{tab:scheduling-comparison} compares these studies that aim to accelerate DL networks using heterogeneous processors.
First, the second column shows the unit of workload distribution used in these studies --- either layer or graph partition. Among them, layer-partition approaches, which partition layers and assign them to heterogeneous processors, incur significant synchronization overhead because they require synchronization at every layer~\cite{mulayer,kang2020scheduling}.

Second, as the third column in \Cref{tab:scheduling-comparison} shows, many studies focus on scheduling only a single DL model~\cite{mulayer,codl,jeong2022tensorrt,collage} or do not perform end-to-end execution experiments for scenarios involving multiple DL models~\cite{kang2020scheduling}, even though various DL models are executed concurrently in practice.

Third, a typical mobile platform offers various configurations for DNN execution; we can select not only the processor (CPU, GPU, or NPU), but also choose the backend implementation (e.g., ONNX Runtime~\cite{onnx-runtime} incorporates its own CPU, XNNPACK~\cite{xnnpack}, and NNAPI~\cite{nnapi} execution providers for the CPU). Additionally, the data types of kernels (32-bit full precision, 16-bit half precision, or 8-bit integer) provide further configuration options, with a trade-off between execution time and output accuracy. While other studies~\cite{kang2020scheduling,band,jeong2022tensorrt,omniboost,codl} often assume a single configuration and focus only on mapping decisions (as shown in the fourth column of \Cref{tab:scheduling-comparison}), we found that no dominant configuration provides the shortest execution time for all types of DNNs; the optimal configuration varies depending on the DNN itself.

Lastly, many studies~\cite{mulayer,kang2020scheduling,codl} estimate the total execution time of a model or subgraph (or set of layers) by summing individual layers' execution times measured offline or predicted, ignoring inter-layer (or graph-level) optimizations. However, these optimizations can significantly affect DL inference performance. Moreover, modern accelerators, such as Qualcomm's NPU, can execute multiple operations concurrently~\cite{qai-sdk}, making the previous estimation method prone to significant errors. We refer to this characteristic of DL inference as the non-linearity of execution time, which is further discussed in \Cref{sec:non-linearity}.

\begin{figure}[t]
    \centering
    \includegraphics[width=0.45\linewidth]{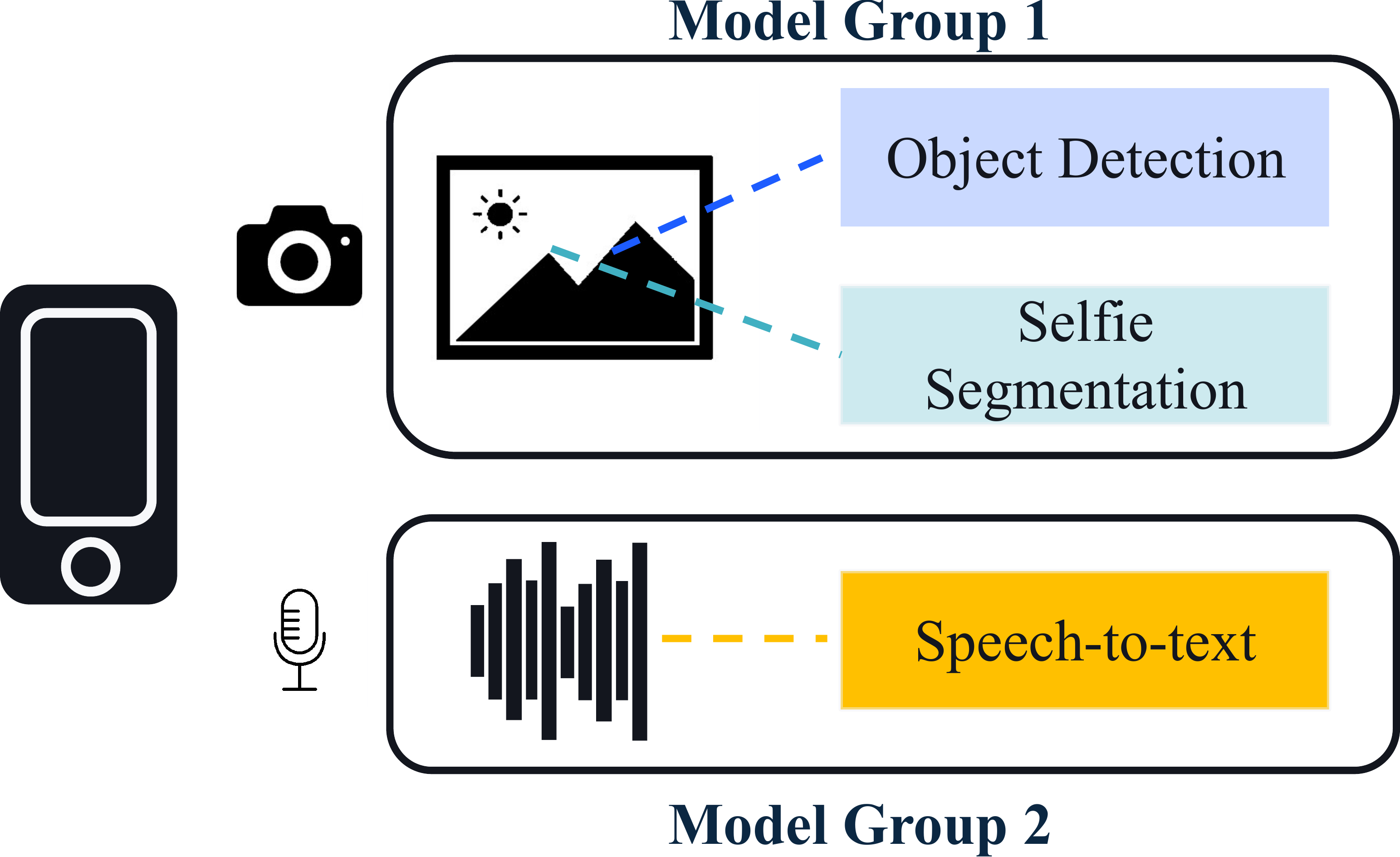}
    \caption{An example of multiple model groups}
    \label{fig:app-example}
\end{figure}

We propose a methodology for scheduling and deploying multiple DL networks on mobile devices equipped with heterogeneous processors. Our approach emphasizes practical considerations and evaluations.
First, observing that DL models on mobile devices often operate periodically based on sensors like cameras or audio sensors, we introduce a \textit{Model Group} --- a set of models that operate based on the same input source in a synchronized manner. In addition, an application may contain multiple model groups. \Cref{fig:app-example} shows an application example that contains multiple model groups: One group may process visual data received from the camera for selfie segmentation~\cite{mediapipe-image-segmenetation}, while another group parses the audio input data to run a speech-to-text pipeline~\cite{whisper2023icml}.

We employ a Genetic Algorithm (GA) to explore the vast search space across multiple dimensions, including graph partitioning, processor mapping, data types, and backend implementations. To effectively explore this complex space, we design three types of chromosomes for graph partition, processor mapping, and priority. Additionally, we address the non-linear execution time by measuring objectives under a user-defined scenario on the target device. To avoid the search from becoming excessively slow, we employ a simple simulator that estimates the objectives with communication cost model and measured execution times of subgraphs. We also develop a runtime equipped with two optimizations, tensor pool and shared buffer, and evaluate GA solutions from various perspective. Based on this methodology, we develop an end-to-end scheduling system for mobile platforms, called \XXX, and systematically evaluate the results.

\section{Background}
\label{sec:background}

\subsection{Performance Characteristics of DNN Inference}
\label{sec:performance-characteristic}


Deep Neural Network (DNN) inference workloads exhibit many unique characteristics~\cite{efficient-processing-dnn,ml-at-fb}, and in this work, we aim to highlight two of them. First, a variety of hardware processors and software solutions are designed to accelerate DNN model execution. While some of them are known for their high performance and are widely adopted across various domains, no single solution is optimal for all cases. We observe that the configuration space is vast, and performance can vary significantly across different configurations.

Next, we observe that predicting or estimating the execution time for a given workload is challenging. The main reasons are inter-layer compiler optimizations and parallel execution on an accelerator, which we call \textit{non-linearity} and detailed discussion can be found in \Cref{sec:non-linearity}. 

\subsubsection{Various Configurations for DNN Execution}

\begin{table*}[th]
    \footnotesize
    \centering
    \caption{Execution time of DNN models on CPU using various configurations available in ONNX Runtime}
    \label{tab:profile_for_various_networks}
    \begin{tabular}{ccccccc}
        \toprule
        \multirow{2}{*}{\textbf{Model}}                                     & \multicolumn{2}{c}{\textbf{Default CPU}}                                                                                      & \multicolumn{2}{c}{\textbf{XNNPACK}} & \multicolumn{2}{c}{\textbf{NNAPI (CPU only)}}                                                         \\
                                                                            & \textbf{fp32}                                                                                                                 & \textbf{fp16}                        & \textbf{fp32}                                 & \textbf{fp16}     & \textbf{fp32}    & \textbf{fp16}  \\
        \toprule
        \textbf{MediaPipe Face Det.}~\cite{mediapipe-face}                  & 2.6   (1.7x)                                                                                                                  & 6.0 (3.8x)                           & \underline{1.6}                               & 5.5 (3.5x)        & 201.0   (128.0x) & 208.5 (132.8x) \\
        \textbf{MediaPipe Selfie Seg.}~\cite{mediapipe-image-segmenetation} & 4.3   (1.4x)                                                                                                                  & 3.5 (1.3x)                           & \underline{3.1}                               & 3.6 (1.2x)        & 106.8   (36.9x)  & 110.2 (36.2x)  \\
        \textbf{MediaPipe Hand Det.}~\cite{mediapipe-hand}                  & 24.3  (4.2x)                                                                                                                  & \underline{5.8}                      & 8.5   (1.5x)                                  & 7.9 (1.4x)        & 198.5   (35.0x)  & 205.1 (35.7x)  \\
        \textbf{MediaPipe Pose Det.}~\cite{mediapipe-pose}                  & 16.3  (2.7x)                                                                                                                  & \underline{6.1}                      & 8.7   (1.4x)                                  & 8.0 (1.3x)        & 286.0   (45.7x)  & 287.7 (47.0x)  \\
        \textbf{TCMonoDepth}~\cite{tc_monodepth}                            & 93.8  (1.3x)                                                                                                                  & \underline{73.2}                     & N/A                                           & N/A               & N/A              & N/A            \\
        \textbf{Fast-SCNN}~\cite{fast_scnn}                                 & 73.2  (2.0x)                                                                                                                  & \underline{37.3}                     & N/A                                           & N/A               & N/A              & N/A            \\
        \textbf{YOLO v8 nano}~\cite{yolov8_ultralytics}                     & 73.0  (1.2x)                                                                                                                  & \underline{58.6}                     & 74.5  (1.3x)                                  & 61.6 (1.1x)       & 638.7   (10.9x)  & 642.9 (11.0x)  \\
        \textbf{MOSAIC (Seg.)}~\cite{weijun2021mosaic}                      & 582.5 (2.7x)                                                                                                                  & 252.6 (1.2x)                         & 373.7 (1.8x)                                  & \underline{213.0} & 1211.7 (5.7x)    & 1208.4 (5.7x)  \\
        \textbf{FastSAM small (Seg.)}~\cite{fastsam}                        & 314.6 (1.6x)                                                                                                                  & 220.3 (6.0x)                         & 297.4 (1.5x)                                  & \underline{192.4} & 1,255.8 (6.5x)   & 1,256.8 (6.5x) \\


        \bottomrule
                                                                            & \multicolumn{6}{r}{\textbf{Device:} Samsung Galaxy S23 Ultra / \textbf{Unit:} ms / \textbf{Aggregation:} Average (100 times)}                                                                                                                                                \\
    \end{tabular}
\end{table*}

DNN Frameworks often allow users to select specific backend implementations and data types. \Cref{tab:profile_for_various_networks} shows various configurations offered by ONNX Runtime (ORT) for CPU. We underline the minimum value among the configurations, and the other columns show the ratio relative to this minimum in parentheses. The result reveals that no single configuration consistently outperforms others across all scenarios.

\paragraph{Backend implementation} Software platform often provides options for choosing which kernel implementations to use. As shown in \Cref{tab:profile_for_various_networks}, ORT provides multiple \textit{Execution Providers} for the CPU: Default CPU, XNNPACK, and NNAPI. The results indicate model performance varies with different execution providers, with some models achieving better performance with the Default CPU, while others run faster with XNNPACK. On the other hand, NNAPI consistently shows the poorest performance across all models. This disparity in performance can be attributed to the distinct kernel implementation and optimizations employed by each execution provider.

\paragraph{Data type} Quantization is a common technique used to improve computational and power efficiency, but using integer kernels can lead to a drop in accuracy. By using 16-bit floating-point (\textit{fp16}), we can avoid this accuracy degradation while accelerating the execution of DNNs, since fp16 covers a similar range as 32-bit floating-point (\textit{fp32}). However, the effectiveness of fp16 depends on model and operation support.

As shown in \Cref{tab:profile_for_various_networks}, we have encountered some counterintuitive cases where executing DNNs with fp16 is slower than executing them with fp32.
This is because, in ORT, if some operators are not supported in fp16, CPU kernels fall back to fp32 kernels, causing extra activation data type conversions and making optimizations such as operation fusion and replacement unavailable. For example, we observed performance degradation in models such as MediaPipe Face Detection.

Since optimal performance settings vary across different scenarios, users should experiment with various DNN kernel configurations, including data type and backend implementations. However, this often requires extensive knowledge of both DL model architectures and the internal mechanisms of DL libraries and kernels.

\begin{table}[th]
    \footnotesize
    \centering
    \caption{Execution time of DNN models on different processors}
    \label{tab:profile_for_various_processors}
    \begin{tabular}{cccc}
        \toprule
        \textbf{Model}                    & \textbf{CPU}  & \textbf{GPU}     & \textbf{NPU}                                   \\
        \toprule
        \textbf{MediaPipe Face Det.}      & 1.6   (5.3x)  & 1.9  (6.3x)      & \underline{0.3}                                \\
        \textbf{MediaPipe Selfie Seg.}    & 3.1   (2.9x)  & 6.5  (6.2x)      & \underline{1.0}                                \\
        \textbf{MediaPipe Hand Det.}      & 5.8   (4.9x)  & 4.9  (4.2x)      & \underline{1.2}                                \\
        \textbf{MediaPipe Pose Det.}      & 6.1   (5.5x)  & 4.9  (4.5x)      & \underline{1.1}                                \\
        \textbf{TCMonoDepth}              & 73.2  (2.3x)  & \underline{31.7} & 32.4 (1.0x)                                    \\
        \textbf{Fast-SCNN}                & 37.3  (2.9x)  & \underline{12.9} & 22.0 (1.7x)                                    \\
        \textbf{YOLO v8 nano (Obj. Det.)} & 58.6  (11.1x) & 16.0 (3.0x)      & \underline{5.3}                                \\
        \textbf{MOSAIC}                   & 213.0 (2.5x)  & \underline{83.8} & 163.9 (2.0x)                                   \\
        \textbf{FastSAM small (Seg.)}     & 192.4 (21.1x) & 43.4 (4.8x)      & \underline{9.1}                                \\

        \bottomrule
        \multicolumn{4}{r}{\textbf{Device:} Galaxy S23 Ultra / \textbf{Unit:} ms / \textbf{Aggregation:} Average (100 times)} \\
    \end{tabular}
\end{table}


\paragraph{Heterogeneous Processors} Different hardware processors such as CPU, GPU, and NPU have various characteristics, making some workloads faster on one processor over others.

\Cref{tab:profile_for_various_processors} provides a comparative analysis of the performance of various DL models across various processors. Similar to CPU, there are multiple configurations for GPU and NPU. We explore all possible combinations but due to space restriction, only present the best configuration (minimum execution time) for each processor. All models are run in fp16. We utilize ORT for CPU and Qualcomm® AI Engine Direct SDK\footnote{Snapdragon and Qualcomm branded products are products of Qualcomm Technologies, Inc. and/or its subsidiaries.}~\cite{qai-sdk} for GPU and NPU.

Six models perform best on the NPU, but performance varies widely; the differences range from 2.9 times to 21.1 times for the CPU, and from 3.0 times to 6.3 times for the GPU. The remaining three models perform better on the GPU, with performance differences compared to the NPU ranging from 1.0 times to 2.0 times. To achieve low latency, running models on the NPU is one of the most easily accessible solutions. However, concurrently running multiple models is not always feasible due to the NPU's computation and memory capacity. Therefore, careful coordination is necessary to utilize all processors effectively, considering the performance differences described above.

\subsubsection{Non-linearity of Execution Times}
\label{sec:non-linearity}

Some scheduling approaches~\cite{kang2020scheduling} estimate the total execution time of a DNN by summing up the execution time of its individual layers. However, this method has limitations due to its reliance on offline execution time measurements, as it fails to account for inter-layer optimizations that vary with graph partitioning explored in this work. Furthermore, modern accelerators can execute multiple operators concurrently, resulting in actual total execution times that significantly differ from the sum of the individual layer execution times. Consequently, previous methods for estimating the total execution time are no longer effective for DNNs optimized with modern compiler techniques and run on modern accelerators.

\begin{table*}[t]
    \setlength{\tabcolsep}{4pt}
    \footnotesize
    \centering
    \caption{Comparison of profiling methods. \textbf{Measured} represents the actual execution time of the network, whereas \textbf{Estimated} predicts the execution time by aggregating the execution time of individual layers. We run this experiment using Qualcomm AI Engine Direct SDK.}
    \label{tab:comparison_bw_and_lw}
    \begin{tabular}{ccccccc}
        \toprule
        \multirow{2}{*}{\textbf{Model}}   & \multicolumn{2}{c}{\textbf{CPU}} & \multicolumn{2}{c}{\textbf{GPU}} & \multicolumn{2}{c}{\textbf{NPU}}                                                               \\
                                          & \textbf{Measured}                & \textbf{Estimated}               & \textbf{Measured}                & \textbf{Estimated} & \textbf{Measured} & \textbf{Estimated} \\
        \toprule

        \textbf{MediaPipe Face Det.}      & 1,611                            & 1,588   (0.99x)                  & 1,914                            & 1,308   (0.68x)    & 305               & 432    (1.42x)     \\
        \textbf{MediaPipe Selfie Seg.}    & 18,616                           & 19,570  (1.05x)                  & 6,461                            & 5,479  (0.85x)     & 1,049             & 2,887  (2.75x)     \\
        \textbf{MediaPipe Hand Det.}      & 15,635                           & 15,754  (1.01x)                  & 4,903                            & 4,052   (0.83x)    & 1,166             & 1,975  (1.69x)     \\
        \textbf{MediaPipe Pose Det.}      & 18,845                           & 18,778  (1.00x)                  & 4,934                            & 3,946   (0.80x)    & 1,107             & 2,182  (1.97x)     \\
        \textbf{TCMonoDepth}              & 206,815                          & 204,095 (0.99x)                  & 31,710                           & 29,142 (0.92x)     & 32,403            & 68,975 (2.13x)     \\
        \textbf{Fast-SCNN}                & 88,425                           & 83,975  (0.95x)                  & 12,852                           & 10,854 (0.84x)     & 21,961            & 62,869 (2.86x)     \\
        \textbf{YOLO v8 nano (Obj. Det.)} & 116,749                          & 117,031 (1.00x)                  & 16,011                           & 14,146  (0.88x)    & 5,260             & 12,635 (2.40x)     \\
        \textbf{MOSAIC}                   & 524,077                          & 509,995 (0.97x)                  & 83,765                           & 78,102 (0.93x)     & 163,879           & 566,094 (3.45x)    \\
        \textbf{FastSAM small (Seg.)}     & 357,265                          & 360,610 (1.01x)                  & 43,413                           & 39.287 (0.90x)     & 9,114             & 15,504 (1.70x)     \\

        \bottomrule
        \multicolumn{7}{r}{\textbf{Device:} Galaxy S23 Ultra / \textbf{Unit:} us / \textbf{Aggregation:} Average (100 times)}                                                                                    \\
    \end{tabular}
\end{table*}

\Cref{tab:comparison_bw_and_lw} shows the measured execution times for various models (\textbf{Measured}) and estimates based on the sum of execution times for all layers in the model (\textbf{Estimated}). The \textbf{Estimated} columns include the relative ratio compared to the measured value in parentheses. The Qualcomm AI Engine Direct SDK provides execution times for fused kernels, attributing most discrepancies in our experiments to parallel execution on modern accelerators and synchronization overheads. For the CPU, the difference between the measured execution time and the estimated one is relatively small, whereas notable discrepancies are observed for both the GPU and NPU. The NPU can execute multiple operations concurrently, which leads to overestimations when the summation method is used. For the GPU, all estimates are smaller than the measured values, likely due to unaccounted overheads such as kernel scheduling and other operational costs. To avoid this non-linearity issue, we employ a \textit{device-in-the-loop profiling} technique, which directly profiles workloads on the real device and evaluates them based on the profiling results.

\subsection{Scheduling Problem Space}

The scheduling problem is well-studied, with numerous studies addressing it. However, it can vary significantly depending on diverse factors, which we enumerate as follows:
\begin{itemize}
    \item \textbf{Number of applications}: For scheduling multiple applications, it is necessary to consider the contention between the applications. Furthermore, when defining the objective, it must be carefully formulated considering all applications. Therefore, as the number of applications increases, the scheduling problem becomes significantly more complex.
    \item \textbf{Timing of scheduling decisions}: Static scheduling involves decisions made at compile time, whereas runtime scheduling makes decisions dynamically during execution.
    \item \textbf{Request pattern}: Incoming user requests can be categorized into two types: \textit{periodic} and \textit{aperiodic}. Periodic requests occur at regular intervals, whereas aperiodic requests arrive unpredictably, typically driven by user behavior or external events.
    \item \textbf{Model group}: DL applications often involve multiple models executing simultaneously with the shared data source. We call such models as a \textit{model group}. For example, vision applications process frames captured by a camera at a fixed frame rate (e.g., 30 FPS).
    \item \textbf{Objectives}: The scheduling problem can have multiple objectives, such as optimizing for speed performance and minimizing energy consumption. Furthermore, speed performance can be measured through two different metrics: throughput and makespan, which is defined as the time to process a single inference request for all models in a model group. We aim to minimize the average and 90th percentile makespans across all model groups.
\end{itemize}

In this paper, we focus on static scheduling of multiple periodic applications to minimize the average and 90th percentile of the makespans for all model groups.

\subsection{Genetic Algorithm}
\label{sec:ga}

Genetic Algorithm (GA) is a meta-heuristic algorithm that solves optimization and search problems by mimicking the process of natural selection. The fittest chromosomes are selected for reproduction to produce offspring of the next generation. The typical steps of GA are as follows:
\begin{itemize}
    \item \textbf{Initialization}: Randomly generate an initial population of chromosomes, which represent potential solutions to the problem. These are typically encoded as strings or arrays.
    \item \textbf{Evaluation}: Each chromosome in the population is evaluated to set its fitness. The fitness indicates the quality of the solution it represents.
    \item \textbf{Selection}: Chromosomes are selected to generate the next population. Generally, those with higher fitness are selected to pass their genes to the next generation.
    \item \textbf{Crossover}: Mate pairs of the selected chromosomes and produce new chromosomes. Parts of their genes are swapped to produce offspring that share genes from both parents.
    \item \textbf{Mutation}: With a given probability, some genes in the new chromosomes are partially modified. This is to introduce variability to the population and avoid converging on suboptimal solutions.
    \item \textbf{Local Search}: With a given probability, explore the solution space near the newly generated chromosome and replace it with a better one.
    \item \textbf{Replacement}: Replace the old population with the new generation of chromosomes.
\end{itemize}



\section{\XXX: Multi-App Scheduler}

We propose \XXX, a multi-application scheduling system that efficiently utilizes heterogeneous processors on a mobile device to serve workloads of multiple deep neural networks. \XXX considers the performance aspects of networks on mobile platforms, as described in \Cref{sec:performance-characteristic}, and adopts a system design that incorporates these characteristics.

\XXX houses a \textit{Static Analyzer} (\Cref{sec:static_analyzer}) component that utilizes a Genetic Algorithm (GA) to explore various configurations for the given networks and finds the optimal configurations for execution. Unlike previous works that make inaccurate estimates of network execution times, the Static Analyzer employs brief on-device measurements for user-defined scenarios. To accelerate the search process, it adopts a simple simulator with a device-in-the-loop profiling technique to measure the on-device execution times of networks and a communication cost regression model.

\XXX can serve multi-application workloads where several networks compete for the same processor. To serve such workloads and satisfy individual application requirements (e.g., real-time latency constraints), we split networks into subgraphs beforehand in the Static Analyzer, and then run subgraphs in the \XXX \textit{Runtime} (\Cref{sec:runtime}) to allow pseudo-preemption~\cite{heimdall} of applications.

Overall, we lay out an extensible system design; \XXX makes use of ``black-box'' subcomponents that can be easily extended to support additional algorithms or frameworks (e.g., new Runtime Engines can be added to support more backend execution frameworks other than ORT and Qualcomm AI Engine Direct SDK).


\begin{figure}[t]
    \centering
    \includegraphics[width=0.8\columnwidth]{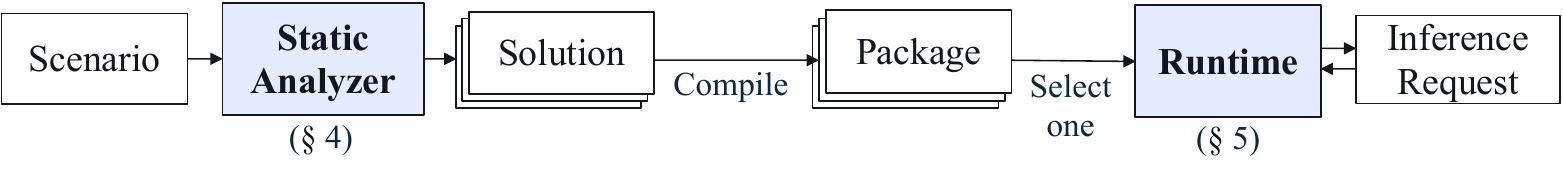}
    \caption{Overall architecture of \XXX.}
    \label{fig:overall-arch}
\end{figure}

\Cref{fig:overall-arch} shows the high-level overview of how \XXX works. It comprises two main components: the Static Analyzer and the Runtime. The Static Analyzer takes user-defined networks and inference scenarios as inputs, and generates feasible solutions. These solutions are then compiled into optimized executable libraries tailored for the assigned processors. Finally, the user selects the most appropriate solution based on the use-case scenario, and submits it to the Runtime, which executes the inference tasks according to the chosen solution.
\section{Static Analyzer}
\label{sec:static_analyzer}

Since scheduling multiple applications with different periods onto heterogeneous processors is an NP-hard problem, most previous works~\cite{kang2020scheduling,band,jeong2022tensorrt,mulayer,codl} rely on heuristic or meta-heuristic algorithms to solve this problem. Following traditional scheduling methodologies, some studies~\cite{kang2020scheduling} typically focus on two main decisions: assigning tasks to specific processors and determining the execution order. However, due to non-linear characteristic of Deep Learning (DL) networks described in \Cref{sec:non-linearity}, an additional crucial decision is needed: how to effectively partition the graph into smaller \textit{subgraphs} that serve as the units of compilation and execution.

\begin{figure}[th]
    \centering
    \subfloat[Partition 1]{\label{subfig:partition1}
        \includegraphics[width=0.4\linewidth]{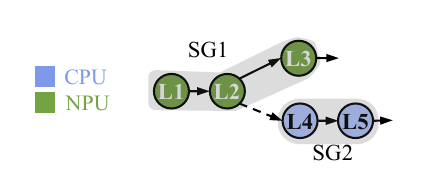}
    }
    \subfloat[Partition 2]{\label{subfig:partition2}
        \includegraphics[width=0.4\linewidth]{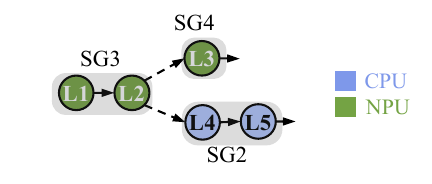}
    }
    \caption{Possible two different partitions with the same mapping: the node color represents the task assigned to processors. For simplicity, we denote layers as L and subgraphs as SG.}
    \label{fig:different-partitions-with-same-mapping}
\end{figure}

\Cref{fig:different-partitions-with-same-mapping} illustrates two distinct scheduling solutions that share the same mapping but employ different partitions. To evaluate these solutions, two key factors must be considered.
First, the non-linearity characteristic discussed in \Cref{sec:non-linearity} impacts performance. Compiling subgraph SG1 as in Figure~\subref*{subfig:partition1}~(layer L1-L3 together) can yield better performance compared to compiling them into two separate subgraphs SG3 and SG4~(L1-L2 and L3, respectively) as in Figure~\subref*{subfig:partition2}. This is because the compiler can optimize the entire subgraph more effectively in the former case, applying inter-layer optimizations, and modern accelerators can leverage inter-layer parallel execution. Second, parallel execution between subgraphs is crucial. In Figure~\subref*{subfig:partition1}, L3 in SG1 cannot be executed in parallel with the layers in SG2 due to the dependency between L2 and L4. In contrast, in Figure~\subref*{subfig:partition2}, SG2 and SG4 do not have data dependency, so it is possible to execute them in parallel. Given these considerations, careful partitioning is essential to achieve optimal performance. To explore these three decisions --- mapping, priority, and partitioning --- simultaneously, we employ a Genetic Algorithm (GA), which has demonstrated state-of-the-art performance in numerous studies.

\begin{figure}[t]
    \centering
    \includegraphics[width=1.0\linewidth]{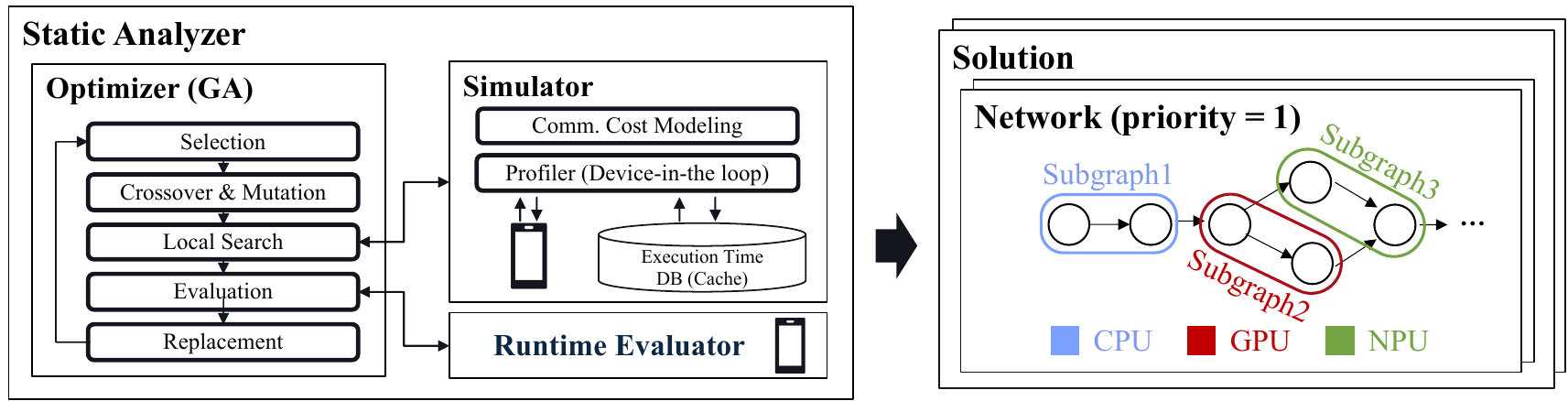}
    \caption{Overall structure of the static analyzer and solution}\label{fig:static-overview}
\end{figure}

\Cref{fig:static-overview} presents the overview of the static analyzer and its solution. The static analyzer consists of three modules: \textit{Optimizer}, \textit{Simulator}, and  \textit{Runtime Evaluator}.

We use the DEAP~\cite{DEAP_JMLR2012} library to implement the Optimizer, which generates new solution candidates. The Static Analyzer employs two evaluation methods: simulation-based evaluation and measurement-based evaluation. To accelerate the local search process, which requires many evaluations, we employ a simple simulator to evaluate candidates. On the other hand, before updating the GA's Pareto solutions, we measure candidates' performance on the target device under a user-defined scenario for more accurate results. The simulator mimics our runtime behavior on the target device, utilizing a communication cost model and a device-in-the-loop profiler, to provide the objectives such as the average and 90th percentile of makespans of model groups. These values serve as the score for assessing the solution. To simulate this behavior, we require the execution time of each subgraph and the communication costs between different processors. The Profiler measures the execution time for each request from the Optimizer on the target device. To determine the backend implementations and data types, we identify the optimal pair for each subgraph and use this as the representative profiling data. For the communication costs, we develop a linear regression model and utilize it during the analysis.

\subsection{Communication Cost Modeling}\label{subsec:communication-cost-modeling}

As a result of the Static Analyzer, DL models are partitioned into subgraphs, and running them on heterogeneous processors incurs communication overhead. Data transfer between processors is facilitated through Remote Procedure Call (RPC) methods, which require a marshalling and unmarshalling process to serialize and deserialize the data, respectively. The overhead of marshalling and unmarshalling is proportional to the data size of the data being transferred. After marshalling, the serialized data is transmitted via high-speed interconnects, such as Network-on-Chip (NoC) and bus systems, which typically operate faster than the main memory, making the main memory the bottleneck in the data transfer.

We model the communication costs by decomposing them into RPC overhead (including the marshalling/unmarshalling) and data transfer time based on the main memory's throughput.

\begin{figure}[th]
    \centering
    \subfloat[Smaller than 1 MiB ]{\label{subfig:smaller-than-1mb}
        \includegraphics[width=.42\linewidth]{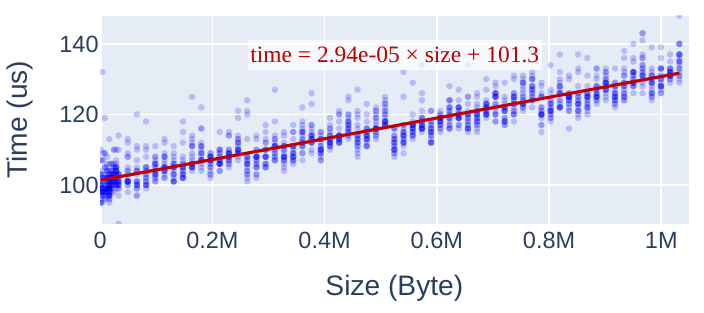}
    }
    \subfloat[Larger than or equal to 1 MiB]{\label{subfig:greater-than-1mb}
        \includegraphics[width=.56\linewidth]{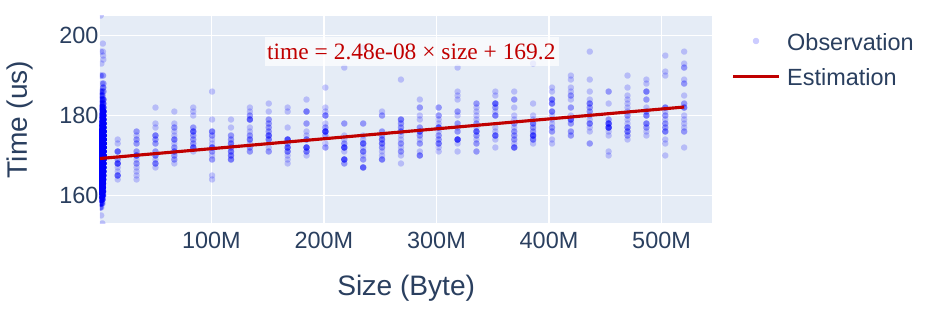}
    }
    \caption{Regression analysis of the RPC overhead}\label{fig:comm-cost-modeling}
\end{figure}

We quantify the RPC overhead by running a microbenchmark and by measuring the time taken to prepare data of varying sizes. We found that the relationship between data size and communication time differs in two regions: smaller than 1MiB and larger than 1MiB. Using piecewise-linear regression, we derive equations that estimate the overhead based on the data size. \Cref{fig:comm-cost-modeling} illustrates the results obtained from the microbenchmark on our target device, Samsung Galaxy S23U.

For memory throughput, we use STREAM benchmark~\cite{McCalpin1995}, which allocates large memory spaces and measures access, copy, and operation times. Through this benchmark, we found the memory bandwidth of Samsung Galaxy S23U, used in our experiments, to be approximately 40 GB/s.



\subsection{Chromosome Design}
A chromosome represents a solution to the problem as described in \Cref{sec:ga}. Since all GA operations are designed based on the chromosome, its design is critical for effective use of GA. In this section, we describe our chromosome design.

\begin{figure}[th]
    \centering
    \includegraphics[width=0.8\linewidth]{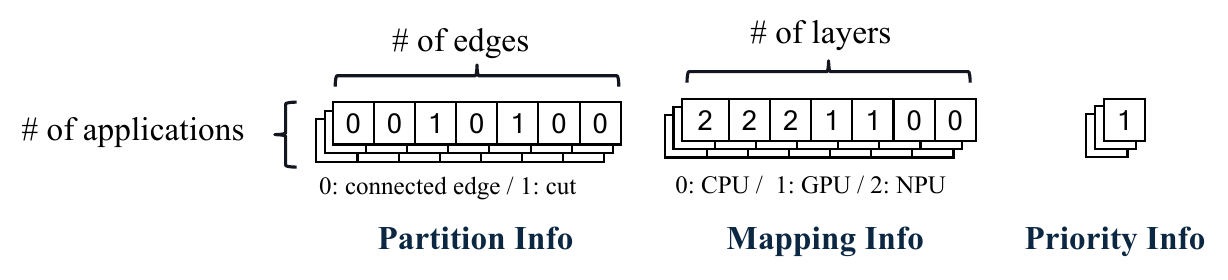}
    \caption{Chromosome design}\label{fig:chromosome-design}

    \subfloat[Partition Information]{\label{fig:partition-example}
        \includegraphics[width=0.45\linewidth]{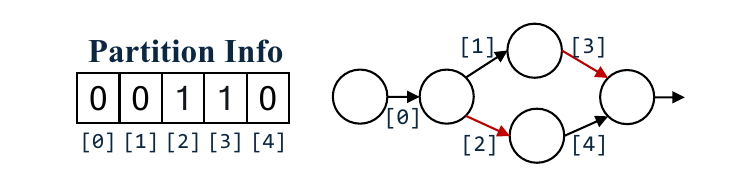}
    }
    \subfloat[Mapping Information]{\label{fig:mapping-example}
        \includegraphics[width=0.45\linewidth]{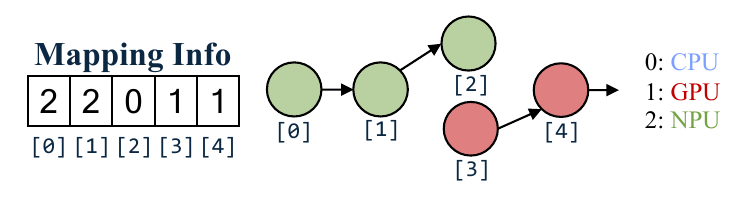}
    }
    \caption{An example of the chromosome and its interpretation}\label{fig:chromosome-example}
\end{figure}

As illustrated in \Cref{fig:chromosome-design}, chromosome in our work consists of partition, mapping, and priority information. The partition information is represented by a set of arrays, one for each network. Each array encodes the edges in the network as binary values, indicating whether each edge is connected (0) or cut (1). Similarly, the mapping information is represented by integer arrays, one for each network, where each array specifies the preferred processor for each layers. The subgraph mapping is determined by a majority vote of the layers within the subgraph. Lastly, the priority information assigns an integer to each network, indicating its precedence in execution.

\Cref{fig:chromosome-example} provides an example to illustrate the interpretation of partition and mapping information. Figure~\subref*{fig:partition-example} shows an array representing the partition information and the original DL network. In this array, \texttt{[2]} and \texttt{[3]} have value 1, meaning the corresponding edges are to be cut (highlighted in red). The remaining edges have value 0, indicating they remain connected. Figure~\subref*{fig:mapping-example} shows the resulting partitioning, which consists of two subgraphs.

Figure~\subref*{fig:mapping-example} shows an array containing mapping information, illustrating how subgraphs are mapped to processors. The green subgraph contains three layers (\texttt{[0]} to \texttt{[2]}) with different processor preferences; the first two layers vote for the NPU, while the third one votes for the CPU. Consequently, the subgraph is assigned to the NPU (green) which receives the majority of votes.


\subsection{GA Process}
\begin{figure}[th]
    \centering
    \includegraphics[width=0.85\linewidth]{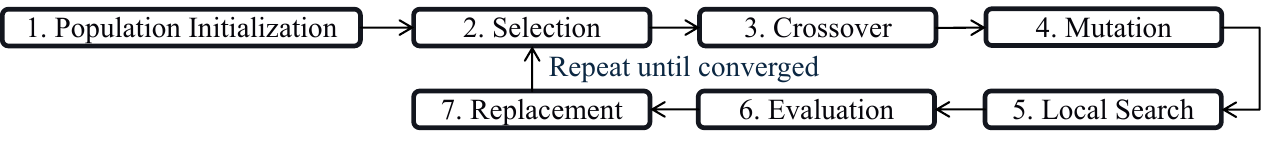}
    \caption{GA process}\label{fig:ga-process}
\end{figure}

\Cref{fig:ga-process} provides an overview of the proposed GA scheduling algorithm. The overall process is similar to typical GAs. It begins with creating an initial population of candidate solutions. To mitigate premature convergence to local optima, we select all candidates as parents for the next generation, instead of selecting only a subset of elite solutions. These parents undergo crossover: one-point crossover is applied to the partition and mapping chromosomes, whereas the Uniform Partially Matched Crossover (UPMX) is used for the priority chromosome. Next, heuristic local search algorithms are applied to the newly generated solutions with a specific probability to explore better nearby solutions. Then, the generated solutions are passed to the Runtime Evaluator to predict on-target performance. Based on these evaluations, the population is updated using the NSGA3 algorithm~\cite{deb2013evolutionary}. This process repeats until the average scores of the population fail to improve for a specific number of iterations, which is set to 3 in this paper.

\subsubsection*{Local Search Algorithm}
We implemented two local search algorithms. The first algorithm merges neighboring subgraphs and updates the solution if it outperforms the current one across all metrics. The other one repositions adjacent layers and applies these changes if they yield a better solution.

\subsubsection*{Solution Assessment}
We employ two methods to evaluate solution candidates. In the evaluation process, we briefly execute the candidates under a user-defined scenario on the target device. 
We incorporate average and 90th percentile of makespans for each model group into optimization objectives, aiming to consider various aspects such as response time and throughput at the same time.

During the local search process, we need to evaluate numerous solutions to compare their performance. To avoid lengthy evaluations, we use a simple simulator, implemented in Python with the SimPy library~\cite{simpy}. This simulator replicates runtime behavior, accounting for both communication and computation costs. Communication costs are calculated using the cost model introduced in \Cref{subsec:communication-cost-modeling}, while computation costs are based on the on-target execution time measured by the Profiler.

\subsubsection*{Device-in-the-loop Profiling}
As discussed in \Cref{sec:non-linearity}, previous works have evaluated scheduling results by measuring execution times for individual operators or layers and summing these values. However, this method lacks accuracy. Therefore, we employed a `Device-in-the-loop Profiling' technique, which involves partitioning models into smaller subgraphs based on the Optimizer's decision, compiling them, and then executing them on the actual device to measure execution time. To reduce the profiling overhead, we stored the profiling results of each subgraph in a database. The Merkle tree algorithm~\cite{merkle1979secrecy} is used to calculate a hash value for each subgraph. This allows us to quickly retrieve and reuse these results, significantly speeding up the profiling process.

\section{Runtime}
\label{sec:runtime}

In this section, we delve into the architecture of \XXX's runtime, exploring its components and how it processes clients' inference requests internally.

\subsection{Overall Architecture}

\begin{figure}[t]
    \centering
    \includegraphics[width=1.0\linewidth]{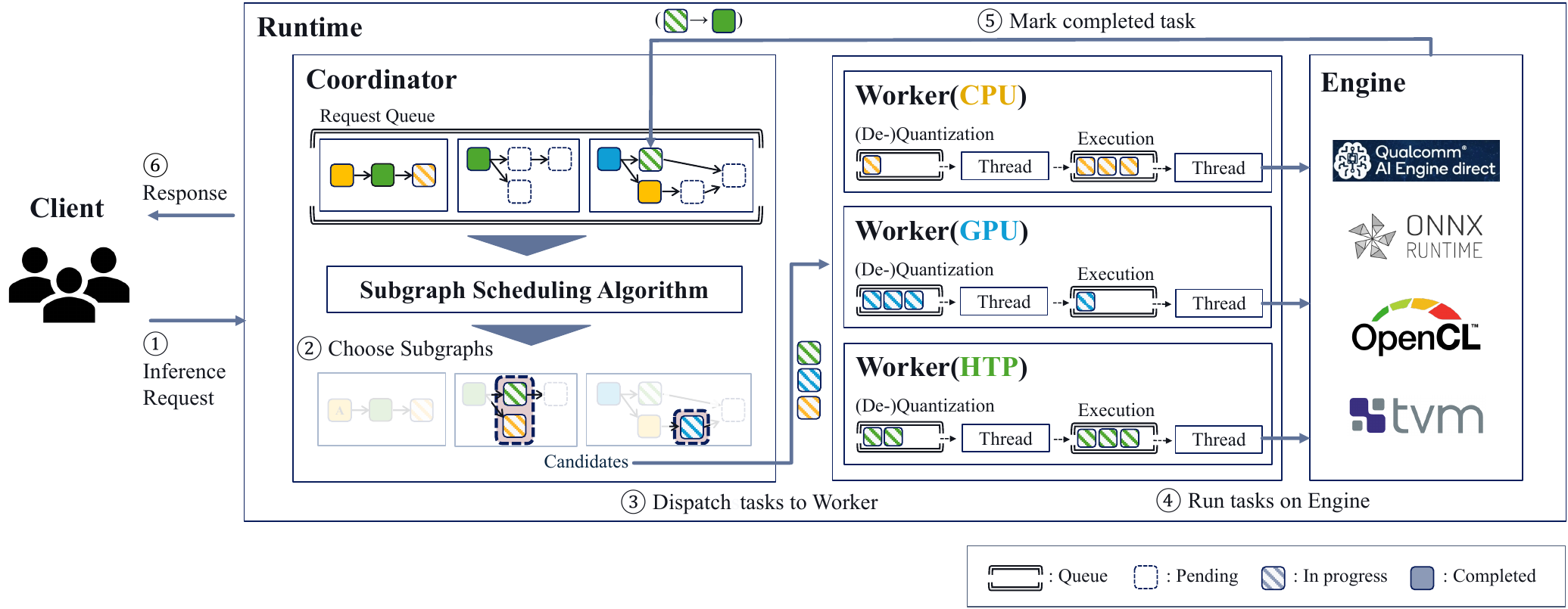}
    \caption{\XXX Runtime architecture and workflow}
    \label{fig:runtime-arch}
\end{figure}

\Cref{fig:runtime-arch} depicts the overall architecture of the Runtime, which comprises three main abstractions: the Coordinator, Worker, and Engine.

The \textit{Coordinator} serves as the external interface to the Runtime, handling inference requests from clients and returning responses. Internally, it manages incoming requests and the entire process to handle them, which is detailed in \Cref{sec:runtime-e2e-workflow}.

Each \textit{Worker} is dedicated to a specific processor such as the CPU, GPU, or NPU. For instance, all CPU subgraph tasks are dispatched to the CPU Worker and then executed sequentially. Before executing tasks, (de-)quantization may be required if the data type of subgraph's input does not match the output of the preceding subgraph. To run task execution and (de-)quantization in parallel, we use two separate threads, each polling items from its dedicated queue.

The \textit{Engine} is a thin abstraction layer over DL runtime frameworks such as Qualcomm AI Engine Direct SDK, ORT, and TVM. It provides a unified interface that hide the details of DL runtime frameworks, enabling users to execute tasks without worrying about framework-specific details. This design also enables \XXX to easily extend to support other DL runtime frameworks.

\subsection{End-to-End Workflow}
\label{sec:runtime-e2e-workflow}

\paragraph{Initialization}
During the initialization of the Runtime, the client registers solutions from the Static Analyzer. The Coordinator initializes the necessary data structures, and Workers load the model libraries embedded in the solution to initialize the corresponding subgraphs, so that all subgraphs are ready to run on the Engines.







\paragraph{Inference Serving}

When a client submits an inference request, it is first queued in the Coordinator's request queue (\circled{1}). The Coordinator thread continuously monitors this queue, searching for schedulable subgraphs with resolved data dependencies (\circled{2}). Upon finding such subgraphs, the Coordinator generates tasks and dispatches them to the corresponding Worker queues (\circled{3}). Each Worker first performs (de-)quantization if needed, followed by task execution using the designated engine (\circled{4}). The results are returned to the Coordinator, which updates the request status (\circled{5}). Once all subgraphs for a request are completed, the final result is sent back to the client (\circled{6}).

\subsection{Optimizations}
\label{subsec:runtime-optimizations}

In \XXX, many inference requests are handled concurrently, and subgraph execution across heterogeneous processors can involve significant overheads for tensor management. We implemented two optimizations, Tensor Pool and Zero-Copy Shared Buffer, to reduce these overheads.

\paragraph{Tensor Pool}
To minimize tensor initialization overheads, we manage a memory pool for tensors, where we initially pre-allocate buffers and reuse them whenever possible.
Once the Runtime has finished using a tensor (e.g., when its corresponding client request is served), the tensor's memory buffer is returned to the pool and retained for further use. Since many inferences of a network involve the same intermediate tensors, recycling memory buffers is effective. When allocating new buffers, we create buffers in chunks of a specific size (2048B in our implementation) to maximize reuse. 
This approach enables a single memory buffer to be used for multiple tensors, even with different sizes.

\paragraph{Zero-Copy Shared Buffer}
\XXX reduces data transfer cost by utilizing zero-copy mechanism~\cite{qai-sdk}, which allocates a \textit{shared buffer} via ION/DMA-BUF memory allocator for Android~\cite{android-ion-allocator,android-ion-to-dmabuf}.

\begin{figure*}[th]
    \centering
    \includegraphics[width=1.0\textwidth]{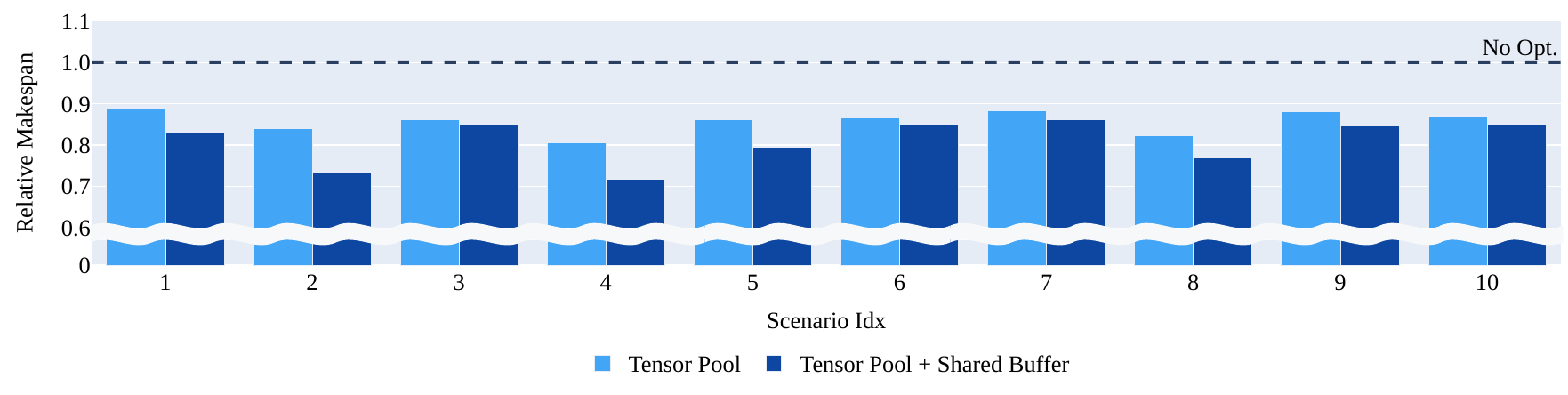}
    \caption{Relative makespan (normalized by the case without any optimization) across the scenarios}
    \label{fig:runtime-makespan}
\end{figure*}

We present an ablation study to analyze the impact of the two optimizations. \Cref{fig:runtime-makespan} shows the makespan ratios for the \XXX solution under the same setup as single model group scenario (\Cref{subsec:eval-single-model-group}), with optimizations applied. Makespan is defined as the duration from a client's inference request to receiving the result, same as in \Cref{subsec:metrics}. 
We evaluate each scenario based on the average makespan, choosing the solution with the smallest maximum makespan.
Applying only the tensor pool improved the makespan by an average of 14.2\%, while enabling the shared buffer further increased the improvement to 18.9\%. When we compute the Pearson correlation coefficients (a measure of linear relationship ranging from -1 to 1) between the makespan reduction with all optimizations enabled and tensor sizes transferred across subgraphs the result is 0.63; this indicates a positive correlation with total tensor size.

\begin{table*}[th]
    \caption{Comparison of time spent on malloc/memcpy/engine execution/free in a single scenario (Mean$\pm{\text{SD}}$)}
    \label{tab:runtime-ablation}
    \footnotesize
    \centering
    \begin{tabular}{ccccccccc}
        \toprule
        \multirow{2}{*}{\makecell[c]{\textbf{Tensor}                                                                                                      \\ \textbf{Pool}}} &
        \multirow{2}{*}{\makecell[c]{\textbf{Shared}                                                                                                      \\ \textbf{Buffer}}} &
        \multicolumn{2}{c}{\textbf{Malloc}}           &
        \multicolumn{1}{c}{\textbf{Memcpy}}           &
        \multicolumn{1}{c}{\textbf{Engine Execution}} &
        \multicolumn{1}{c}{\textbf{Free}}                                                                                                                 \\
                                                      &   &
        \makecell{\textbf{Time}                                                                                                                           \\ \textbf{(ms)}} & \makecell{\textbf{\# of} \\ \textbf{Alloc}} &
        \makecell{\textbf{Time}                                                                                                                           \\ \textbf{(ms)}} &
        \makecell{\textbf{Time}                                                                                                                           \\ \textbf{(ms)}} &
        \makecell{\textbf{Time}                                                                                                                           \\ \textbf{(ms)}} \\
        \midrule
        X                                             & X & 14.2 \(\pm{0.6}\) & 1734 & 965.4 \(\pm{37.5}\)  & 8999.1 \(\pm{175.9}\) & 311.2 \(\pm{12.6}\) \\
        O                                             & X & 3.3 \(\pm{0.1}\)  & 17   & 329.1  \(\pm{20.8}\) & 8653.7 \(\pm{125.8}\) & 1.9 \(\pm{0.1}\)    \\
        O                                             & O & 4.4 \(\pm{0.5}\)  & 17   & 283.7  \(\pm{16.9}\) & 8201.3 \(\pm{56.5}\)  & 1.5 \(\pm{0.1}\)    \\
        \bottomrule
    \end{tabular}
\end{table*}

In \Cref{tab:runtime-ablation}, we compare the malloc, memcpy (occurs only on input tensor copy), engine execution, and free times for Scenario 5 with the baseline without optimization. The tensor pool method yields significant reductions: malloc time by 76.8\% and free time by 99.4\% through buffer reuse. In contrast, the baseline incurs higher memcpy time due to memory allocation overheads, such as physical memory allocations and page table updates, which occur during memory access rather than during malloc. By reusing tensors, the tensor pool method decreases memcpy time by 65.9\%. Additionally, the engine execution time is reduced by 3.8\%.

When utilizing the shared buffer, we observe a slight increase in malloc time due to RPC memory allocation overheads. However, this impact is minimal as malloc time constitutes only a negligible portion of the total execution time. The shared buffer offers additional benefits, including a 70.6\% reduction in memcpy time, which is a 4.7 percentage points improvement over the tensor pool method alone. Furthermore, the engine execution time is reduced by 8.9\% with the shared buffer, outperforming the tensor pool method by 5.1 percentage points, primarily by minimizing input data transfer times of the input data from preceding subgraphs.

\section{Evaluation}
\label{sec:eval}


In this section, we comprehensively evaluate \XXX using randomly generated scenarios with nine state-of-the-art models. We first describe our experimental setup in \Cref{subsec:experimental-setup}, including target device, randomly generated scenarios, and baselines. Evaluation metrics are presented in \Cref{subsec:metrics}, followed by results for single and multiple model groups in \Cref{subsec:eval-single-model-group,subsec:eval-multi-model-group}, respectively. 

\subsection{Experimental Setup}
\label{subsec:experimental-setup}

In the experiments, we use Samsung Galaxy S23 Ultra, equipped with a Qualcomm Snapdragon® 8 Gen 2 SoC, which features an 8-core CPU, a Qualcomm® Hexagon™ GPU, and a Qualcomm® Adreno™ NPU. We have implemented \XXX using Python and C/C++. The Static Analyzer is written in Python (9.8K LoC), and the Runtime is written in C/C++ (4.3K LoC). Although \XXX currently depends on Qualcomm AI Direct SDK version 2.29.0 and ONNX Runtime libraries version 1.20.1, it can be easily extended to support other libraries.

To evaluate the performance of our scheduling approach, we introduce two baselines based on heuristic methods: \textit{NPU Only} and \textit{Best Mapping}. \textit{NPU Only} runs all models solely on the NPU, which is highly optimized for neural network inference and generally offers the best performance among processors for most models in our evaluation. In contrast, \textit{Best Mapping} is a search-based heuristic that profiles each model on each processor and then finds Pareto mappings by adjusting the mappings based on execution times. This approach considers interactions among all networks but does not incorporate subgraph partitioning.

\begin{figure}[th]
    \vspace{-2em}
    \begin{minipage}{0.5\textwidth}
        \centering
        \footnotesize
        \setlength{\tabcolsep}{2pt}
        \captionof{table}{DL models used in experiments}
        \begin{tabular}{clccc}
            \toprule
            \textbf{Idx} & \textbf{Model}                                             & \textbf{\# MACs} & \textbf{\# Params} \\
            \toprule
            1            & MediaPipe Face Det.~\cite{mediapipe-face}                  & 39.2 M           & 0.6 M              \\
            2            & MediaPipe Selfie Seg.~\cite{mediapipe-image-segmenetation} & 72.3 M           & 0.1 M              \\
            3            & MediaPipe Hand Det.~\cite{mediapipe-hand}                  & 410.8 M          & 2.0 M              \\
            4            & MediaPipe Pose Det.~\cite{mediapipe-pose}                  & 444.2 M          & 3.4 M              \\
            5            & TCMonoDepth~\cite{tc_monodepth}                            & 2313.2 M         & 0.2 M              \\
            6            & Fast-SCNN~\cite{fast_scnn}                                 & 2358.9 M         & 1.1 M              \\
            7            & YOLO v8 nano~\cite{yolov8_ultralytics}                     & 4891.3 M         & 3.2 M              \\
            8            & MOSAIC (Seg.)~\cite{weijun2021mosaic}                      & 22055.1 M        & 1.8 M              \\
            9            & FastSAM small (Seg.)~\cite{fastsam}                        & 22325.1 M        & 11.8 M             \\
            \bottomrule
        \end{tabular}
        \label{tab:models}
    \end{minipage}\hfill
    \begin{minipage}{0.5\linewidth}
        \centering
        \begin{figure}[H]
            \centering
            \subfloat[Single MG]{\label{fig:single-mg-grid}
                \includegraphics[width=0.45\linewidth]{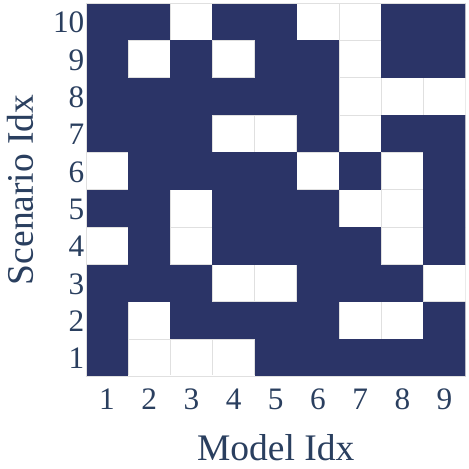}
            }
            \subfloat[Multi MG]{\label{fig:multi-mg-grid}
                \includegraphics[width=0.45\linewidth]{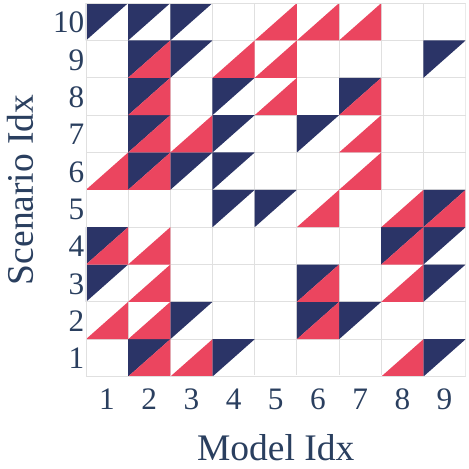}
            }
            \caption{Scenario Configurations. Filled box at $(x, y)$ implies model $x$ is included in scenario $y$. Different colors imply involvement in different model groups.}\label{fig:scenario}
        \end{figure}
    \end{minipage}
    \vspace{-1em}
\end{figure}

\Cref{tab:models} lists the models used across various application domains. To evaluate our scheduling approach, we create synthetic scenarios by randomly selecting from the nine models. These scenarios are categorized into two types: single model group and multiple model groups. \Cref{fig:scenario} illustrates how each scenario is composed.

\begin{itemize}
    \item \textbf{Single model group:} We create ten scenarios, each containing six randomly selected models.
    \item \textbf{Multiple model groups:} For each scenario, we form two model groups of three models, maintaining the same settings as in the single model group experiments.
\end{itemize}

We assume that all DL model groups run periodically, aligned with the inputs from sensors in mobile devices such as cameras and audio sensors, which operate at fixed frequencies. The period of the model groups is a crucial factor in defining the scheduling problem; excessively long periods can trivialize scheduling, whereas overly short periods can make solutions unfeasible. To provide a comprehensive comparison, we evaluate performance across various periods.


For a scenario \(\mathcal{S}\) consisting of model groups \({G_{1}, G_{2}, \cdots, G_{N}}\), we define the \textit{base period} \(\Bar{\phi}_{G_{i}}\) for each model group \(G_{i}\) as follows:
\[ \Bar{\phi}_{G_{i}} = \sum_{m \in G_{i}} \left(\min_{p \in P}\tau_{p}(m) \right) \cdot N \cdot (1 + \epsilon) \]

where \(\tau_{p}(m)\) denotes the execution time of model \(m\) on processor \(p\), and P represents the available processors. We calculate the sum of the profiled execution times of all models included in the group, on the fastest processor. Multiplying by N and \(1 + \epsilon\) provides slack for scheduling among multiple model groups, where N is the number of model groups and \(\epsilon\) is a small constant (0.1 in our experiments).

\[ \Phi_{(\alpha, G_{i})} = \alpha \cdot \Bar{\phi}_{G_{i}} \]

We then adjust the period \(\Phi_{(\alpha, G_{i})}\) by multiplying a coefficient called \textit{period multiplier} $\alpha$ with the base period. A smaller \(\alpha\) indicates stricter Service Level Objective (SLO), whereas a larger \(\alpha\) relaxes the deadline, makes it easier to meet the SLO~\cite{band}.

\subsection{Metrics}
\label{subsec:metrics}

To evaluate the performance of multiple model groups across different methods and period settings, we adopt the score definitions from XRBench~\cite{kwon2023xrbench}, designed to assess real-time multi-task multi-model (MTMM) workloads. XRBench provides performance evaluations at various levels, including per inference requests, per model, per use case scenario, offering a comprehensive assessment from diverse perspectives. Aggregating scores of individual model groups is challenging due to the complexity introduced by multiple models and groups; however, XRBench offers a method to combine these various scores into a normalized range [0, 1], enabling fair comparison across different configurations.

Since XRBench score already accounts for multiple models, we can use it without major changes. However, to incorporate the concept of model groups, we introduce an additional term and make slight modifications to the intermediate scores.

\textit{Makespan} measures the time required to process a single inference request, including all models in the model group. For the \(j\)-th request of model group \(G_{i}\), the makespan \(\Theta_{G_{i}}^{(j)}\) is computed as:

\[ \Theta_{G_{i}}^{(j)} = \max_{m \in {G_{i}}}({T_{f}}^{(j)}(m)) - \min_{m \in {G_{i}}}({T_{s}}^{(j)}(m))\]
where \({T_{s}}^{(j)}(m)\) and \({T_{f}}^{(j)}(m)\) are the starting and finishing timestamps, respectively, for the \(j\)-th request of model \(m\).

\textit{QoE Score} is measuring the quality of experience, being calculated as the ratio of successfully processed requests within the deadline (equal to the period in our setup) to the total number of requests made to the model group $G_i$ (denoted as $J_{G_i}$).

\[ QoE Score(\alpha, G_i) = \frac {|\{j | \Theta_{G_{i}}^{(j)} \leq  \Phi_{(\alpha, G_{i})}\}|}{J_{G_i}}  (1 \leq j \leq J_{G_i}) \]

\textit{Realtime Score} measures how quickly inference requests are handled. The parameter $k$ controls the sensitivity to the deadline, and we use $k=15$, the same value as the original paper.
\[ RtScore^{(j)}{(\alpha, G_{i})} = 1 \big/ (1 + e^{k \cdot (\Theta_{G_i}^{(j)} - \Phi_{(\alpha, G_{i})} )}) \]

Since we do not alter the original model's architecture or parameters when partitioning models in \XXX, we omit \textit{Accuracy Score} and assume it is always 1.0. 
We also ignore the \textit{energy score} because considering energy consumption is beyond the scope of this paper. Extending \XXX to cover energy consumption is left for future work.

Using the above scores, the final score of a scenario at a multiplier $\alpha$ is defined as
\[
    Score(\alpha, \mathcal{S}) =
    \frac{1}{N} \cdot
    \sum_{G_{i}}
    \bigg(
    \frac{
        \sum_{j}{RtScore^{(j)}{(\alpha, G_{i})}}}
    {J_{G_{i}}
    }
    \cdot QoEScore(\alpha, G_i)
    \bigg)
\]
We can evaluate the performance of a scenario under different period settings, by changing the multiplier ($\alpha$) values.

To compare the performance between \XXX and other baselines, it is useful to determine how much load each method can handle without performance degradation (e.g., frame drop). To this end, we define the \textit{saturation multiplier} as the minimum period multiplier that achieves the maximum XRBench score of 1.0:

\[
    \alpha_{\mathcal{S}}^{*} = min \{ \, \alpha \mid Score(\alpha, \mathcal{S}) = 1.0 \, \}
\]

In cases where multiple solutions emerge due to multiple objectives, we employ the median score value of these solutions to determine the saturation multiplier.

\subsection{Single Model Group}
\label{subsec:eval-single-model-group}


\begin{figure}[th]
    \includegraphics[width=.95\textwidth]{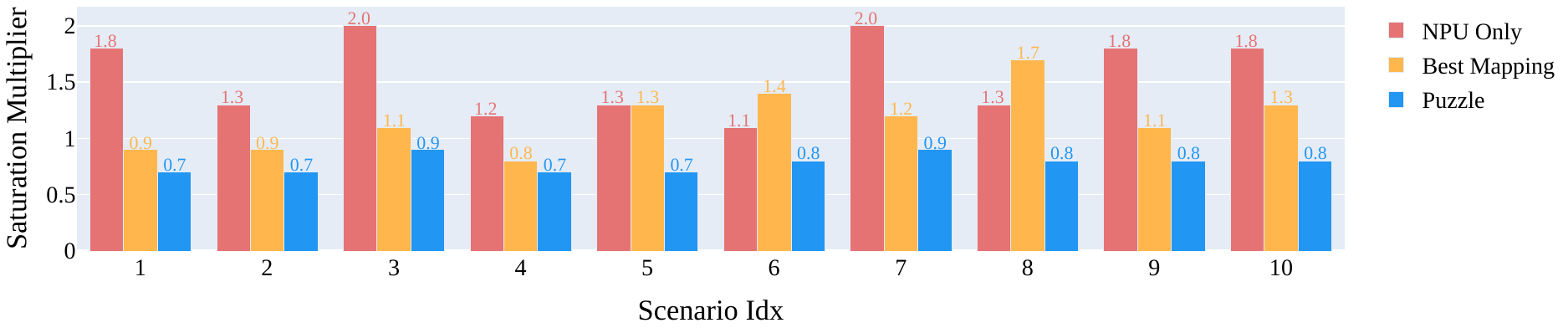}
    \caption{Performance comparison under different scenarios in single model group experiments}
    \label{fig:single-mg-gain}
\end{figure}

We evaluate the scheduling performance of \XXX against two baseline approaches: Best Mapping and NPU Only. This evaluation covers ten scenarios, with all solutions obtained by searching at a period multiplier 1.0.

\Cref{fig:single-mg-gain} examines how the saturation multiplier (defined in \Cref{subsec:metrics}) varies across all scenarios under different scheduling methods.
When the period multiplier decreases, it indicates a higher request frequency (shorter period), and when the period multiplier increases, it signifies a lower request frequency (longer period).
\XXX demonstrates notable resilience, achieving a saturation multiplier ($\alpha_{\mathcal{S}}^{*}$) of $0.78\pm{0.08}$~(Mean$\pm{\text{SD}}$). This means it can handle request frequencies higher than those encountered during the search process. This robustness likely stems from \XXX's fine-grained model partitioning during exploration, which offers a performance buffer compared to the coarser approach of Best Mapping. In contrast, Best mapping has a saturation multiplier of $1.17\pm{0.27}$, indicating that while it accounts for model interactions, its lack of partitioning limits its optimization potential. Meanwhile, the NPU Only method, with a saturation multiplier of $1.56\pm{0.35}$, highlights the drawbacks of relying solely on a single high-performance processor.

\begin{figure}[th]
    \centering
    \subfloat[Scenario 1]{
        \label{fig:single-mg-individual-scenario-0}
        \includegraphics[width=.5\textwidth]{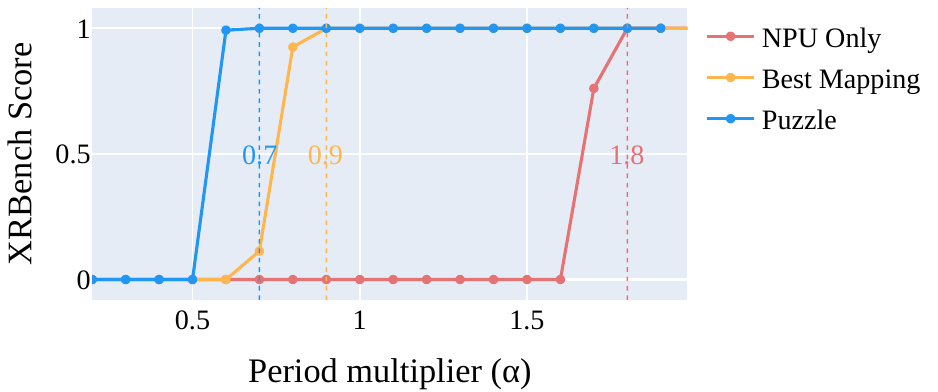}
    }
    \subfloat[Scenario 8]{
        \label{fig:single-mg-individual-scenario-7}
        \includegraphics[width=.5\textwidth]{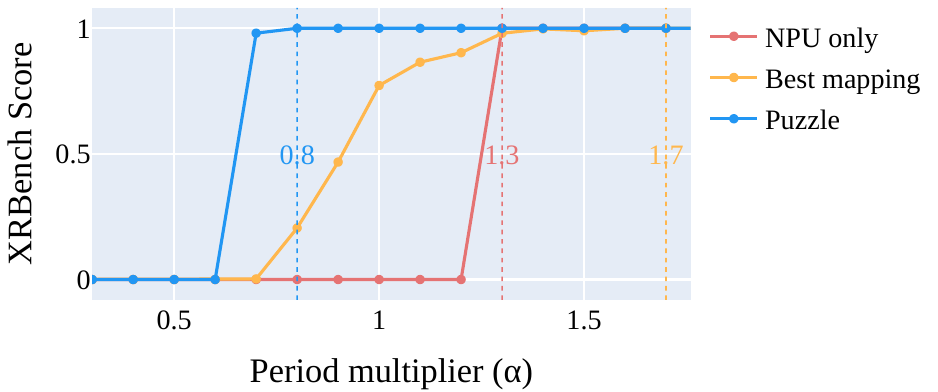}
    }
    \caption{Comparison of two scenarios with a single model group, varying multipliers}
    \label{fig:single-mg-individual-scenario}
\end{figure}

\Cref{fig:single-mg-individual-scenario} shows the XRBench score trends for \XXX and the two baselines as a function of the period multiplier in two scenarios. This analysis indirectly evaluates the robustness of each method under varying periods. In Scenario 1, performance aligns with earlier observations. However, Scenario 8 reveals an exception: Best Mapping’s score reaches 1.0 at a higher multiplier than NPU Only. 
This discrepancy arises because Best Mapping relies solely on model profiling, neglecting potential contention for shared resources, such as main memory, NoC, bus, and the CPU. The CPU, which handles tasks like scheduling, job dispatching, and other system operations, experiences significant fluctuations. As the system approaches saturation, instability increases, resulting in variable performance.
For example, executing Best Mapping's solution ten times at a period multiplier of 1.0 yielded scores ranging from 0.64 to 0.9. Near the saturation point (multiplier 1.3), scores fluctuated between 0.9 and 1.0, demonstrating the effect of resource contention on performance. In contrast, \XXX exhibits a transient state at a single period multiplier value of 0.6, with its scores ranging narrowly from 0.98 to 1.0.

\subsection{Multi Model Groups}
\label{subsec:eval-multi-model-group}

In this section, we compare \XXX with two baselines across scenarios for multiple model groups.

Unlike the single model group case, multiple model groups compete for processors to optimize their respective objectives, which results in performance trade-offs among them.

\begin{figure}[th]
    \centering
    \subfloat[$\alpha = 1.4$]{
        \label{fig:multi-mg-pareto-1_4}
        \includegraphics[width=.4\textwidth]{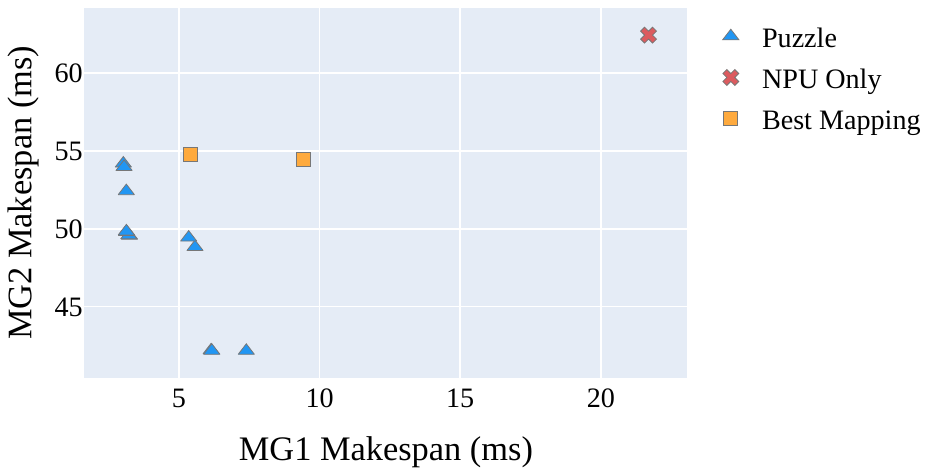}
    }
    \subfloat[$\alpha = 0.9$]{
        \label{fig:multi-mg-pareto-0_9}
        \includegraphics[width=.4\textwidth]{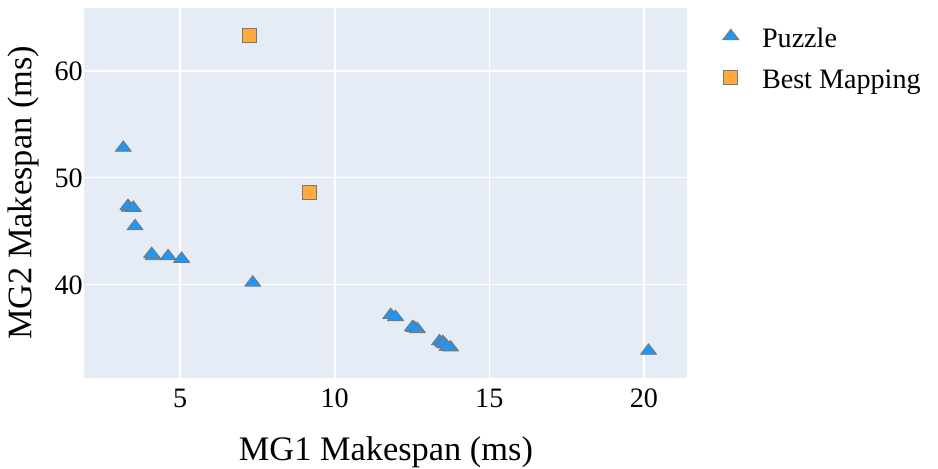}
    }
    \caption{Makespan distribution of solutions of Scenario 10 in two different period settings ($\alpha=0.9, 1.4$).}
    \label{fig:multi-mg-pareto}
\end{figure}

To understand this trade-off, we present the average makespan of each model group in Scenario 10, as shown in \Cref{fig:multi-mg-pareto}. For long periods (lenient workloads), Figure~\subref*{fig:multi-mg-pareto-1_4} indicates that \XXX consistently outperforms other methods. Figure~\subref*{fig:multi-mg-pareto-0_9} presents the results of tight periods, where \XXX also demonstrates superior performance. To facilitate comparison, we omit results from the NPU Only baseline due to its exponential increase in makespan values and system failure caused by accumulated inference tasks. In addition, \XXX provides various Pareto solutions, allowing users to select a scheduling solution according to their needs.

\begin{figure}[th]
    \centering
    \includegraphics[width=.95\textwidth]{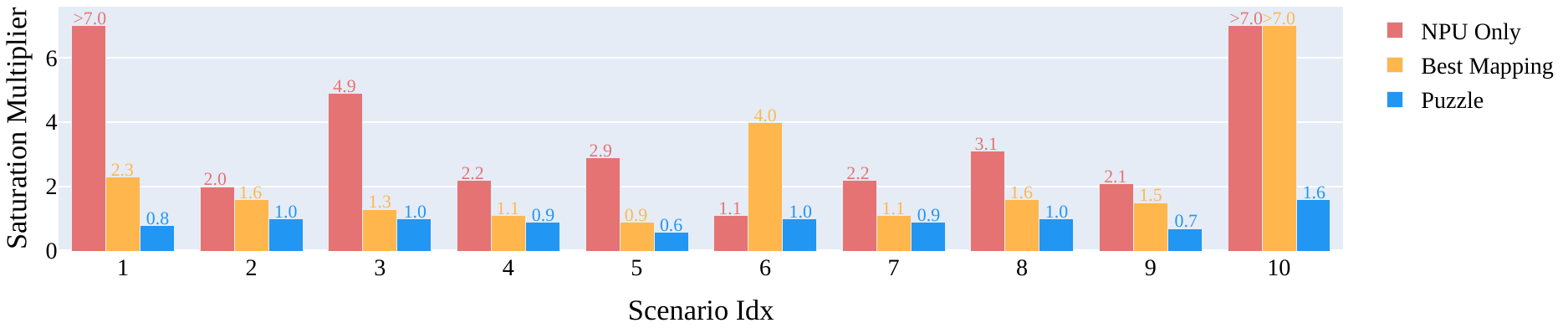}
    \caption{Performance gain under different scenarios in multi model group experiments.}
    \label{fig:multi-mg-gain}
\end{figure}

\Cref{fig:multi-mg-gain} summarizes results across all ten scenarios. Following the single model group experiments format, we report saturation multiplier ($\alpha_{\mathcal{S}}^{*}$). \XXX achieves a score of $0.95 \pm{0.27}$, performing similarly to the single model group scenario and outperforms other baselines in all scenarios. In contrast, Best Mapping and NPU Only scores $2.24 \pm{1.90}$ and $3.45 \pm{2.12}$, respectively. Although the total number of models are the same as the single model group scenario, the baselines exhibit greater performance degradation, which can be attributed to two main reasons. First, the Best Mapping does not account for shared resource contention, as we highlighted in the single model group experiments. Second, all baselines rely on a coarse-grained, model-level mapping. In a non-preemptive system, executing a large model can monopolize the processor, preventing other models from running until it completes. Consequently, this leads to other tasks missing their deadlines. To address these challenges, \XXX accounts for the resource contention during exploration and optimizes scheduling decisions to minimize overall overhead including this contention.

\begin{figure}[th]
    \centering

    \subfloat[Scenario 6]{
        \label{fig:multi-mg-individual-scenario-5}
        \includegraphics[width=.45\textwidth]{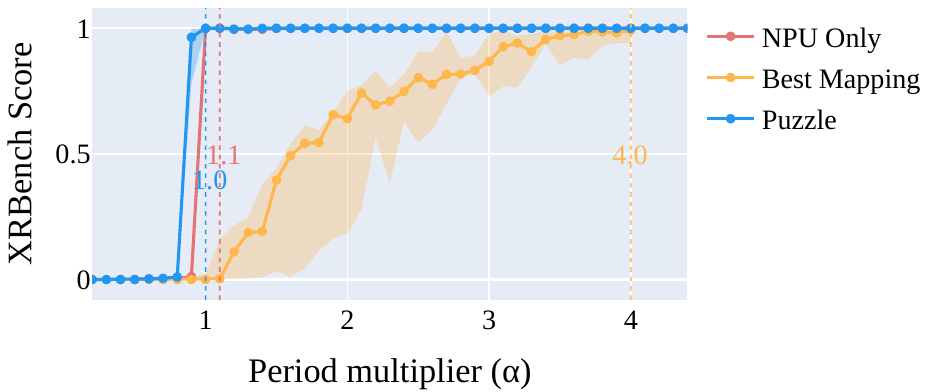}
    }
    \subfloat[Scenario 10]{
        \label{fig:multi-mg-individual-scenario-9}
        \includegraphics[width=.45\textwidth]{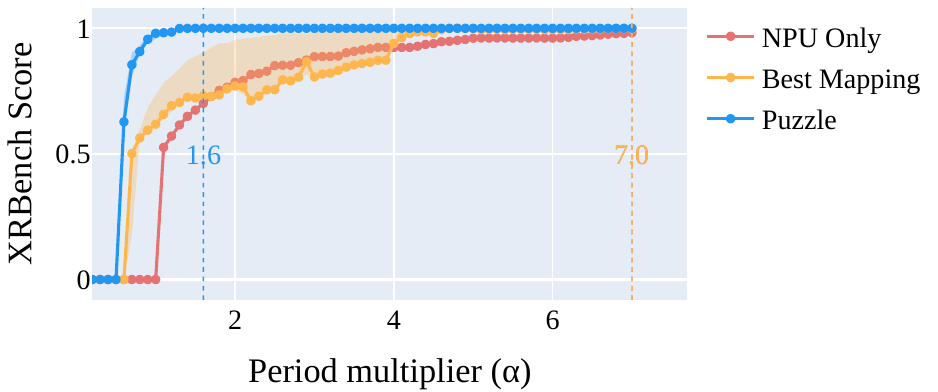}
    }
    \caption{Comparison of a scenario with multiple model groups, varying multipliers}
    \label{fig:multi-mg-individual-scenario}
\end{figure}

In \Cref{fig:multi-mg-individual-scenario}, We further investigate the distinct behavior observed in Scenarios 6 and 10 by analyzing the XRBench score as a function of the period multiplier. In the multi-model group setting, both \XXX and Best Mapping generate multiple solutions. For each multiplier, we visualize the min/max scores across these solutions as shaded areas and the median score as a solid line.

Scenario 6 consists of five MediaPipe models and one YOLOv8 model, exhibiting distinct performance characteristics. MediaPipe models are computationally lightweight, executing in 0.3ms to 1.2ms on the NPU and around 5-6 times slower on other processors (1.9ms to 6.6ms). In contrast, YOLOv8 is computationally heavy, taking 5ms on the NPU, 17ms on the GPU, and 60ms on the CPU. Since all models perform best on the NPU, \XXX generates solutions that prioritize NPU allocation, showing similar performance to the NPU Only baseline. Best Mapping attempts to optimize resource utilization by running some of MediaPipe models on the CPU, expecting performance consistent with profiling results, while fixing YOLOv8 to the NPU due to its size. However, CPU performance fluctuations, as mentioned earlier, cause diverse and unstable solutions in Best Mapping, preventing it from meeting the 1.0 XRBench score. Notably, \XXX
also explored these solutions but they were not selected due to the performance fluctuations identified through brief execution evaluation.

In Scenario 10, \XXX significantly outperforms both baselines. This scenario includes two model groups: one with three lightweight models (Mediapipe series) and another with three computationally heavy model (YOLOv8, FastSCNN, and TCMonoDepth). Best Mapping attempts to minimize makespans by co-locating heavy and small models to the same processor, but this leads to starvation for the lightweight models during heavy model execution. As a result, the baselines fail to achieve a score of 1.0, leading to higher saturation multipliers. In contrast, \XXX utilizes pseudo-preemption by dividing models into smaller subgraphs, enabling it to reach a score of 1.0 with a relatively low period multiplier.

\section{Related Work}




\paragraph{Model Partitioning}
As hardware technology advances, many scheduling systems have emerged to leverage heterogeneous processors. A straightforward approach is to partition the network and parallelize its execution. $\mu$-layer~\cite{mulayer}, CoDL~\cite{codl}, and Lin et al.~\cite{lin2023online} split operators (e.g., along the channel dimension) and distribute them across different processors. By utilizing multiple processors, these systems achieve latency speedup with better energy efficiency. Collage~\cite{collage} and Jeong et al.~\cite{jeong2022tensorrt} further consider partitioning graphs into subgraphs, which reduces communication overhead and facilitates graph-level optimizations.


While the above systems focus on scheduling single-model neural network inference, some focus on optimizing the concurrent execution of multiple neural network models. OmniBoost~\cite{omniboost} provides automatic mapping by using Monte Carlo Tree Search with reinforcement learning and a simple CNN model to estimate processor throughput.
Band~\cite{band} divides networks into subgraphs based on operator availability and uses dynamic programming to decide which subgraphs to run.

Kang et al.~\cite{kang2020scheduling}, the most similar work to \XXX, utilize a Genetic Algorithm (GA) to determine mapping. However, their method derives model partition directly from the mapping decision without exploration. In contrast, our approach simultaneously explores model partition, mapping, and priority using GA, leading to more optimal solutions, as discussed in \Cref{sec:static_analyzer}.


\paragraph{Performance Estimation Method}
When analyzing execution of deep learning applications on mobile devices, we can consider two approaches: model-based and profiling-based. The former relies on analytical or ML-based models that capture the computation and communication characteristics~\cite{qi2017paleo,neuralpower,codl,nn-meter}. While this approach works well in general, training a high-quality model requires significant effort. Within profiling-based methods, Kang et al~\cite{kang2020scheduling} estimate execution time based on layer-wise measurements, which we demonstrate to be inaccurate in \Cref{sec:non-linearity}.
\XXX adopts a hybrid approach. For communication costs, \XXX utilizes simple regression models. It employs a device-in-the-loop profiler to measure subgraph execution times. To enhance accuracy, it evaluates candidates by briefly executing them on the device at the end of each GA generation.

\paragraph{Static vs. Dynamic}
Static scheduling systems, including \XXX, make scheduling decisions at compile time or design time through static analysis~\cite{kang2020scheduling}. This approach enables exploration of a broad search space and provides minimal runtime overheads since all decisions are pre-determined. On the other hand, dynamic scheduling systems~\cite{band} make scheduling decisions in real-time, making them suitable for adapting to the system status. Moreira et al.~\cite{moreira2007scheduling} propose integrating static and dynamic scheduling, combining the benefits of static and dynamic scheduling, which presents an intriguing avenue for further research.
\section{Conclusion}
As on-device AI technology advances, scheduling multiple Deep Learning (DL) models on mobile devices becomes increasingly challenging. We have identified the challenges in serving multiple DL models: diverse hardware/software configurations and the lack of a universal solution. Although NPUs are integrated into mobile devices to accelerate DL networks, their performance capabilities alone are insufficient to handle the growing demand for concurrent DL models. To address these challenges, we propose a novel framework called \XXX, which schedule DL models onto the heterogeneous processors embedded within mobile devices to accelerate multiple DL models.

\XXX consists of a Static Analyzer and a Runtime component. The Static Analyzer uses a Genetic Algorithm to schedule multiple DL models at design time. It explores partitioning decisions that divide DL models into subgraphs, as well as the mapping and priority (or execution order) decisions. To address the non-linearity characteristic of DL identified in this paper, it evaluates candidates with brief execution and device-in-the-loop profiling. It also investigates configurations involving various kernel implementations during exploration. The Runtime employs tensor pool and zero-copy techniques to minimize system overhead from data transfer and memory allocation.

Through extensive experiments with random scenarios, we show that \XXX can support 3.7 and 2.2 times higher request frequency on average compared to the two heuristic baselines, NPU Only and Best Mapping, respectively, while satisfying the equivalent level of real-time requirements.

\bibliographystyle{ACM-Reference-Format}
\bibliography{software}

\end{document}